\documentclass[journal]{IEEEtran}

\usepackage{amsmath,amssymb,amsfonts}

\usepackage{graphicx}
\usepackage{booktabs}
\usepackage{subcaption}
\usepackage{hyperref}
\usepackage{microtype}
\usepackage{xurl}

\hypersetup{colorlinks=true,linkcolor=blue,citecolor=blue,urlcolor=blue}
\begin{document}

\title{The Field Knows: Cross-Dimensional Geometry from Navigation to Black Holes}

\author{Chenghao~Xu
\thanks{C. Xu is with the School of Artificial Intelligence and Robotics and National Engineering Research Center of Robot Visual Perception and Control Technology, Hunan University, Changsha, Hunan, China. Email: chenghaoxu@hnu.edu.cn.}}

\maketitle

\begin{abstract}
We introduce a continuous metric field framework trained by a single causal contrastive loss. The framework encodes a scene into coefficients of a fixed symmetric matrix basis, assembles them into a Lie algebra element, and exponentiates the result to a Riemannian or Lorentzian metric. Across dimensions, this field discovers the full spectrum of geometric structures: from obstacle-avoiding geodesics in robot navigation across planar and manipulator configuration spaces, to event horizons of black holes in Lorentzian spacetime. Extensive zero-shot generalization studies demonstrate that the field captures transferable geometric structure rather than memorizing specific configurations. In the black hole setting, the causal loss spontaneously evolves genuine black-hole-like structures with the correct Lorentzian signature. The same loss, the same architecture, and the same training protocol produce the full range of geometric phenomena across dimensions. The field knows geometry, and geometry knows physics.
\end{abstract}

\begin{IEEEkeywords}
Continuous metric field, basis matrix parameterization, causal contrastive loss, black hole horizon emergence, Cartan spectral clamp, zero-shot generalization, cross-dimensional geometry
\end{IEEEkeywords}

\section{Introduction}

Planning through geometry, rather than through policy, is the central idea of spatial intelligence. In prior work~\cite{Xu2025space}, this was realized through discrete geometric generators: each obstacle contributed a localized term to the metric tensor, and these contributions were composed to form a global metric. The approach worked, with perfect zero-shot generalization from single-scene training, but the metric existed only where generators were placed. The geometry was a collection of discrete contributions, not a continuous field.

In this paper, we deepen spatial intelligence into a continuous field. Every point in space carries its own metric, produced by an encoder that maps the entire scene to coefficients of a fixed symmetric matrix basis. There are no per-obstacle generators, no Router. A single causal contrastive loss shapes the entire field: collision-free paths are cheap, obstacle-penetrating paths are expensive. The geometry is no longer a sum of discrete parts; it is a unified object that exists everywhere.

We push this continuous field across dimensions. In planar and 6-degree-of-freedom (6-DOF) manipulator configuration spaces, we conduct systematic zero-shot generalization studies to characterize how the field transfers from training to unseen configurations. The results show that the field captures transferable geometric structure rather than memorizing specific scenes. In Lorentzian spacetime, we replace the obstacle constraint with a causal one (falling in is cheaper than escaping) and ask what geometry emerges. The same framework, trained by the same loss, produces obstacle-avoiding geodesics in robot spaces and event horizons in spacetime.

We present four dimensional regimes:
\begin{itemize}
    \item \textbf{2D planar}: A continuous metric field for point robot navigation, with zero-shot generalization across systematically varied test suites.
    \item \textbf{6D manipulator}: A 6-DOF arm in a cluttered workspace, where the configuration-space metric generalizes from single-scene training to fully random environments.
    \item \textbf{3D Lorentzian}: Ba\~{n}ados-Teitelboim-Zanelli (BTZ)-like black hole emergence, where the metric spontaneously develops an event horizon under the causal constraint.
    \item \textbf{4D Lorentzian}: Schwarzschild-like black hole emergence with suppressed cross-terms and near-spherical symmetry.
\end{itemize}

Our contributions are:
\begin{enumerate}
    \item A continuous metric field framework that deepens spatial intelligence from discrete generators to a unified geometric object spanning every point in space, and extends it across dimensions from robot navigation to black hole horizons, all under a single causal contrastive loss.
    \item Systematic zero-shot generalization studies in both planar and 6-DOF manipulator spaces, demonstrating that the continuous field captures transferable geometric structure rather than memorizing specific configurations.
    \item Demonstration that the causal loss spontaneously evolves genuine black-hole-like structures with the correct Lorentzian signature in 3D and 4D spacetime, including the identification of a mechanism that prevents trivial collapse and forces genuine horizon formation.
\end{enumerate}

The remainder of this paper is organized as follows. Section~\ref{sec:related} situates the paper relative to procedural decision systems, metric-based geometric modeling, and our prior discrete-generator framework. Section~\ref{sec:arch} describes the continuous metric field architecture, including the basis matrix parameterization, the causal contrastive loss, and the Cartan spectral clamp. Section~\ref{sec:2d} presents the 2D planar navigation experiments and zero-shot generalization results. Section~\ref{sec:6d} extends the framework to 6-DOF manipulator configuration spaces. Section~\ref{sec:3d} and Section~\ref{sec:4d} present the emergence of black hole horizons in 3D and 4D Lorentzian spacetime. Section~\ref{sec:shell_flip} characterizes the double-horizon failure mode. Section~\ref{sec:conclusion} concludes.

\section{Related Work}
\label{sec:related}

This paper is best understood in relation to two prior lines of work. The first treats geometry as an auxiliary ingredient inside a larger decision procedure. The second gives geometry and metric structure a more central role in motion generation and control. Our work is closest to the second line, but ultimately departs from both. It does not use geometry merely to support planning, nor does it treat geometry only as a structure for composing policies. Instead, it takes the continuous metric field itself as the primary learned object.

\subsection{Geometry as Auxiliary Structure}

Most existing systems in robotics and learning still place the main problem-solving burden on an external procedure. In potential-field and trajectory-optimization methods, geometry enters as cost, force, constraint, smoothness prior, or local guidance inside a planner or optimizer~\cite{Khatib1986Potential,Ratliff2009CHOMP,Kalakrishnan2011STOMP}. Even when these quantities are continuous and spatially structured, they are introduced to bias, regularize, or accelerate an explicit solver rather than to define the primary object of learning. In more recent learned systems, neural networks predict paths, heuristics, or arrival-time-like fields that make downstream search or planning more effective~\cite{Qureshi2021MPNet,Ni2023NTFields}. Here too the learned quantity remains closer to a planning aid than to a geometric object in its own right: it shapes the agent's decision process, but the burden of producing behavior still lies with an external planner, optimizer, or search rule. These works are important, but in all of them geometry remains auxiliary: it supports the decision process rather than becoming the primary learned object.

\subsection{Geometry- and Metric-Centered Motion Generation}

A smaller but more relevant line of work moves closer to our viewpoint by elevating geometry and metric structure themselves. Riemannian Motion Policies and RMPflow introduce task-dependent metrics with clear geometric meaning and use them to combine local behaviors consistently~\cite{Ratliff2018RMP,Cheng2021RMPflow}. Related work on learned Riemannian manifolds and dynamic optimization fabrics likewise shifts attention toward geometry-aware or geometry-induced motion generation rather than purely external search~\cite{BeikMohammadi2023Riemannian,Spahn2023Fabrics}. This already marks a substantial shift away from agent-centered planning and control, because geometry is no longer just a penalty term or heuristic attached to an explicit solver; it becomes part of the representation through which motion is organized and composed. In that sense, these works help establish that geometry can play a central computational role rather than merely supporting an agent-side decision process.

At the same time, the endpoint of those frameworks is still a composed policy or control law. The metric determines how local motion policies are weighted, transformed, or fused, but it is not itself the final learned object from which the full solution geometry is read out. Our prior work made a clearer break from the agent-centered view by treating intelligence as residing in geometry itself and by learning scene-conditioned Riemannian metrics through discrete geometric generators and routing~\cite{Xu2025space}. The present paper is motivated by the limitations of that discrete construction: the metric should no longer be assembled from localized slots, but should instead become a continuous field defined at every point, with a single assembly principle that extends across dimensions from Riemannian robot spaces to Lorentzian spacetime.

\section{The Continuous Metric Field}
\label{sec:arch}

This section describes the continuous metric field framework. The metric is generated in three stages: a scene encoder produces basis coefficients at each spatial point, these coefficients are assembled into a Lie algebra element via a fixed symmetric matrix basis, and the result is exponentiated to a Riemannian or Lorentzian metric. A single causal contrastive loss shapes the entire field, with a structural constraint that prevents the loss from being trivially satisfied. We also describe the Cartan spectral clamp, which provides principled control over the metric's curvature and, in the black hole setting, enables the mechanism that forces genuine horizon formation.

\subsection{Architecture}

The framework produces a continuous metric field $G(x) \in \mathrm{SPD}(n)$ (symmetric positive definite $n\times n$ matrices) from a scene representation, , where $n$ is the dimension of the manifold (2 for 2D, 6 for 6D, etc.). The architecture has three stages:

\textbf{Stage 1: Scene Encoding.} The scene is discretized into a grid and encoded by a lightweight backbone network. For 2D and 6D, where the input is a spatial density field with obstacles, we use a convolutional neural network (CNN): a 2D CNN for planar scenes, and a 3D CNN for the 3D obstacles which are projected to the 6D joint configuration space. For 3D and 4D spacetime, the input is a fixed random latent grid, encoded by a 3D and 4D CNN, respectively, serving as a generative prior for the metric field. At any spatial query point $x$, the feature map is linearly interpolated to produce a feature vector. This vector, augmented with a positional encoding of $x$, is then passed to an MLP for the next stage.

\textbf{Stage 2: Basis Matrix Assembly.} The MLP outputs $m = \dim\mathrm{Sym}(n) = n(n+1)/2$ scalar coefficients per spatial point, , where $\mathrm{Sym}(n)$ is the space of symmetric $n \times n$ matrices. These coefficients, denoted $t_k(x)$, are the coordinates of $H(x)$ in a fixed basis $\{X_k\}_{k=0}^{m-1} \subset \mathrm{Sym}(n)$. This forms a Lie algebra element $H(x) \in \mathfrak{gl}(n)$ (the Lie algebra of $n\times n$ matrices):
\begin{equation}
    H(x) = \sum_{k=0}^{m-1} t_k(x) \cdot X_k,  \quad X_k \in \mathrm{Sym}(n).
    \label{eq:H_basis}
\end{equation}
$H(x)$ is a symmetric matrix in the Lie algebra, which is then exponentiated to produce the metric. Unlike the prior work~\cite{Xu2025space} which relied on a Router to select among generator slots, the present framework has no Router. Every coefficient is produced directly by the encoder. The choice of basis $\{X_k\}$ is a critical architectural decision, as it determines the algebraic coupling between the metric components.

The basis $\{X_k\}$ follows a unified construction for any dimension $n$. For $m = n(n+1)/2$ basis matrices:
\begin{itemize}
    \item $X_0 = I$: the identity matrix, controlling the isotropic (trace) component.
    \item $n-1$ traceless diagonal matrices of the form $X_k = \mathrm{diag}(\underbrace{1,\ldots,1}_{k}, -k, 0,\ldots,0)$ for $k=1,\ldots,n-1$, controlling anisotropic deformations along independent axes.
    \item $n(n-1)/2$ symmetric off-diagonal matrices, each with entries $1$ at positions $(i,j)$ and $(j,i)$ and $0$ elsewhere, controlling shear between dimensions.
\end{itemize}
This construction generalizes across all dimensions, yielding $m = n(n+1)/2$ basis matrices for any $n$ (e.g., $m=3$ for 2D, $m=6$ for 3D, $m=10$ for 4D, and $m=21$ for 6D). The diagonal degrees of freedom are fully decoupled at the Lie algebra level, providing independent control over each metric component.

\textbf{Stage 3: Metric Assembly.} Before exponentiation, the Lie algebra element $H(x)$ passes through a Cartan spectral clamp to prevent numerical overflow:
\begin{equation}
    \tilde{H}(x) = c(x) \cdot H(x), \quad
    c(x) = \frac{\tanh(\|H(x)\| / L)}{\|H(x)\| / L + \varepsilon} > 0,
    \label{eq:cartan}
\end{equation}
where $L$ is the clamp scale (a hyperparameter controlling the maximum allowed spectral radius), $\|H(x)\|$ denotes the spectral norm (the largest absolute eigenvalue of $H(x)$), and $\varepsilon = 10^{-12}$ prevents division by zero. Since $c(x) > 0$, $\tilde{H}(x)$ is a positive scalar multiple of $H(x)$ whose spectral radius does not exceed $L$. From $\tilde{H}(x)$, the group element $E(x) = \exp(\tilde{H}(x)) \in \mathrm{GL}(n)$ (the general linear group of invertible $n\times n$ matrices) is obtained via the matrix exponential. The metric field is then assembled in a unified form across all dimensions:
\begin{equation}
    G(x) = E(x)^\top \eta\, E(x), \quad E(x) = \exp(\tilde{H}(x)),
    \label{eq:G_unified}
\end{equation}
where $\eta$ is the metric signature matrix, a fixed diagonal matrix of $\pm 1$ entries that determines whether the geometry is Riemannian or Lorentzian. For Riemannian geometry (2D and 6D obstacle avoidance), $\eta = I$ and $G = E^2 = \exp(2\tilde{H})$. For Lorentzian geometry (3D and 4D black hole), $\eta = \mathrm{diag}(-1, +1, \ldots, +1)$. 

By Sylvester's law of inertia, $\mathrm{sig}(G) = \mathrm{sig}(\eta)$ as long as $E(x)$ is non-singular (guaranteed by the matrix exponential), where $\mathrm{sig}(\cdot)$ denotes the signature of a matrix (the numbers of positive and negative eigenvalues). For Riemannian metrics (2D and 6D), a final minimum eigenvalue shift of $\max(0, 0.05 - \lambda_{\min})$ is applied to prevent numerical instability when eigenvalues approach zero, where $\lambda_{\min}$ is the smallest eigenvalue of $G(x)$; this shift is omitted for Lorentzian metrics (3D and 4D) since the negative eigenvalue of the time component is physically meaningful and must not be altered.

\subsection{Causal Contrastive Loss and the $t_0>0$ Structural Constraint}
\label{sec:loss_and_constraint}

The entire field is trained by a single causal contrastive loss:
\begin{equation}
    \mathcal{L} = \frac{\overline{\mathcal{L}}_{\text{pos}}}{\overline{\mathcal{L}}_{\text{neg}} + 0.1} \;+\; 0.001 \cdot \overline{\mathcal{L}}_{\text{pos}},
    \label{eq:loss}
\end{equation}
where path costs are computed as discrete Riemannian (or Lorentzian) lengths:
\begin{equation}
    \mathcal{L}_{\text{path}} = \sum_i \sqrt{|\delta_i^\top G(m_i) \delta_i|},
    \label{eq:pathcost}
\end{equation}
with $\delta_i = p_{i+1} - p_i$ as the displacement vectors and $m_i = (p_i + p_{i+1})/2$ as the segment midpoints, where $p_i$ are the discrete points along the path.

The loss has two terms. The first is the ratio $\overline{\mathcal{L}}_{\text{pos}} / (\overline{\mathcal{L}}_{\text{neg}} + 0.1)$, which drives the field to separate positive and negative path costs. The second is the regularization term $0.001 \cdot \overline{\mathcal{L}}_{\text{pos}}$, which serves as an anchor. It provides a gentle downward pressure on the absolute magnitude of $\mathcal{L}_{\text{pos}}$, preventing both costs from drifting upward together while maintaining a small ratio. The coefficient $0.001$ is chosen small enough that the ratio term dominates the gradient direction; the regularization merely prevents unbounded growth.

\begin{figure*}[t]
	\centering
	\begin{subfigure}[b]{0.24\textwidth}
		\centering
		\includegraphics[width=\textwidth]{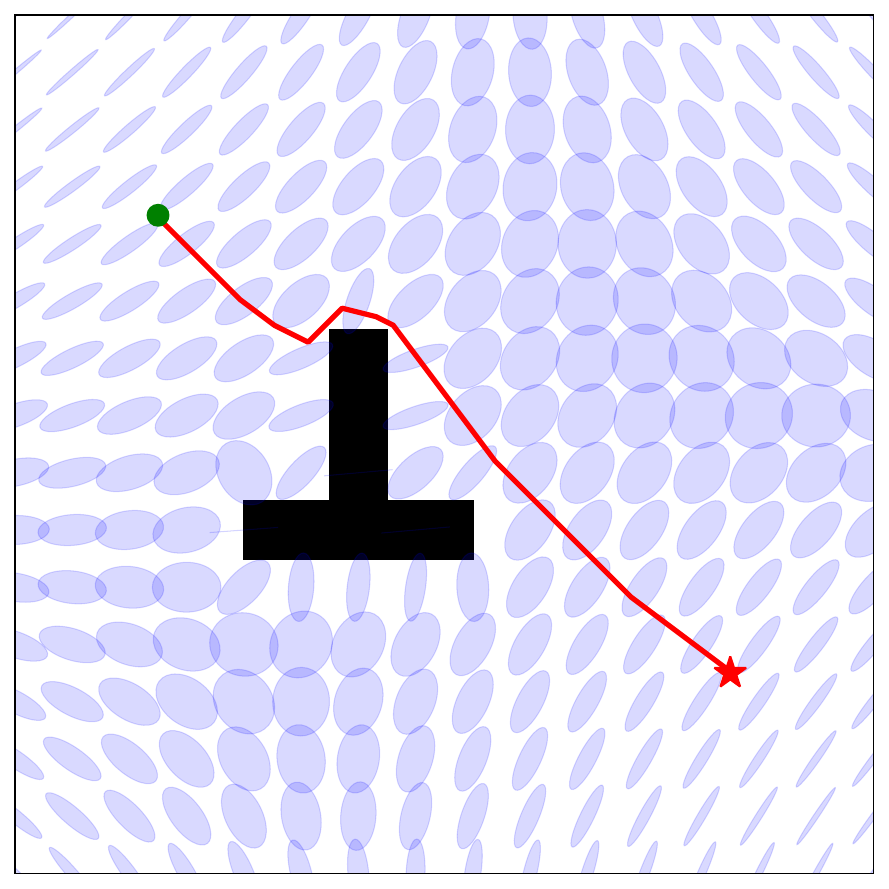}
		\caption{L-shaped corridor}
	\end{subfigure}\hfill
	\begin{subfigure}[b]{0.24\textwidth}
		\centering
		\includegraphics[width=\textwidth]{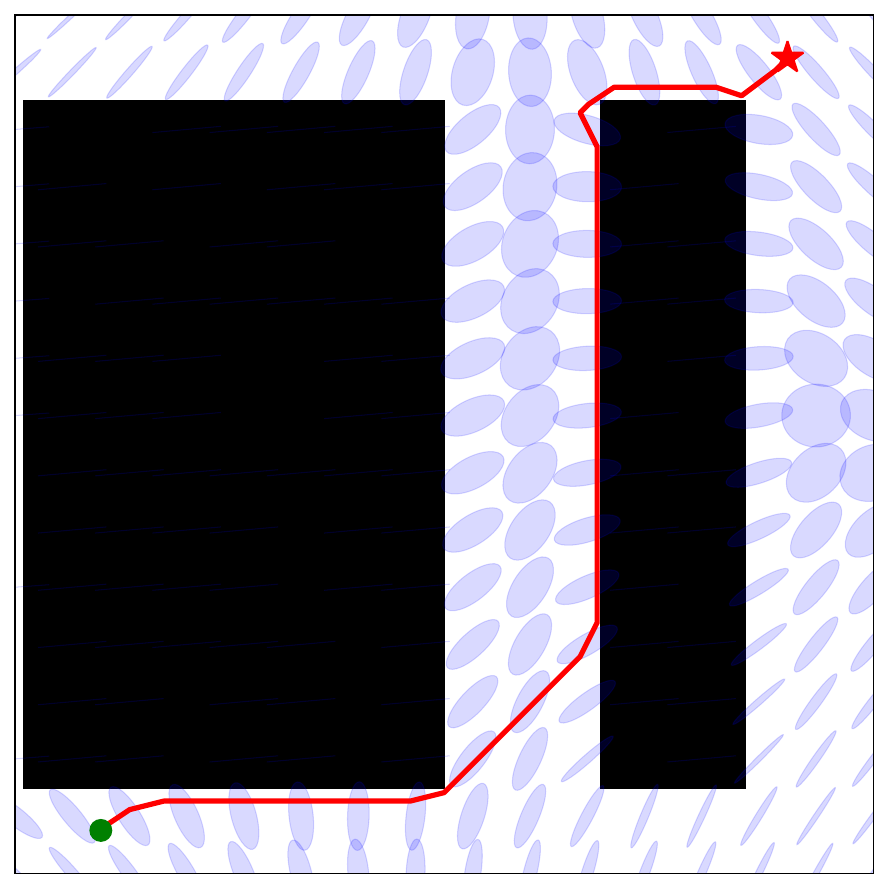}
		\caption{Narrow passage}
	\end{subfigure}\hfill
	\begin{subfigure}[b]{0.24\textwidth}
		\centering
		\includegraphics[width=\textwidth]{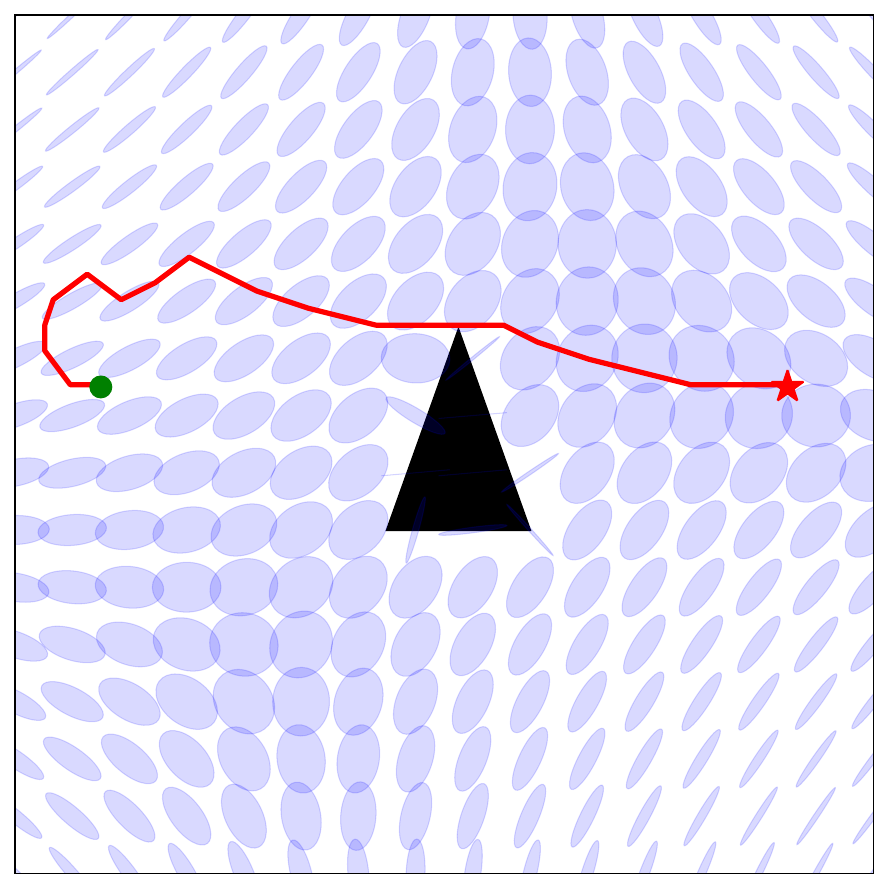}
		\caption{Triangle obstacle}
	\end{subfigure}\hfill
	\begin{subfigure}[b]{0.24\textwidth}
		\centering
		\includegraphics[width=\textwidth]{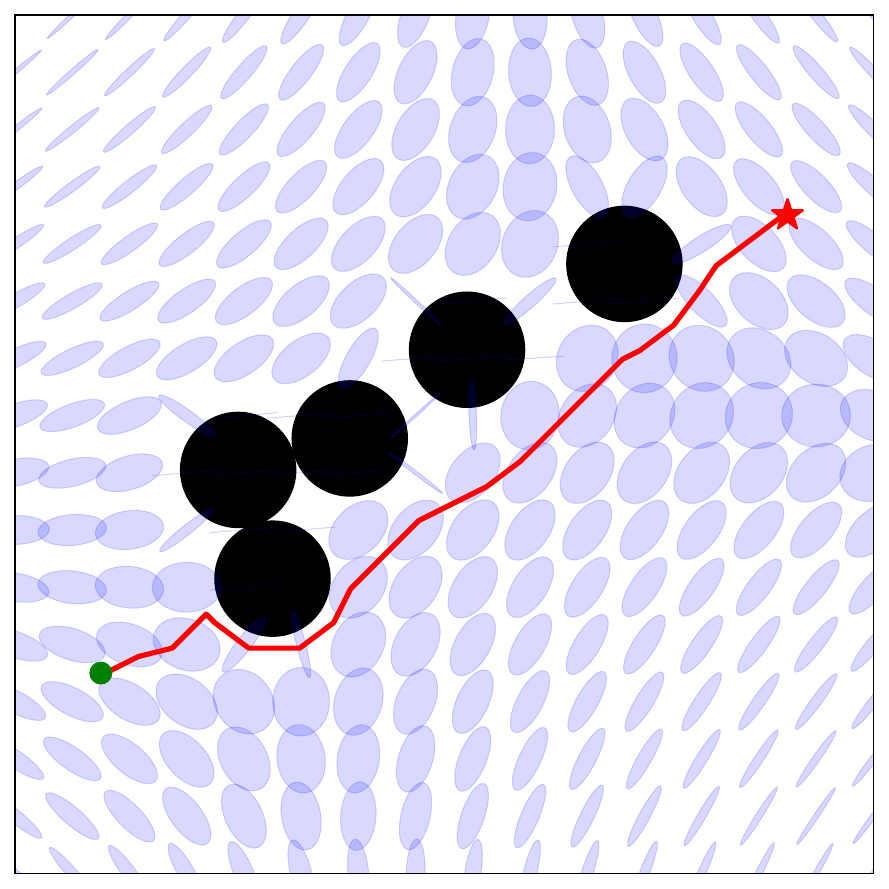}
		\caption{Random obstacles}
	\end{subfigure}
	\caption{Zero-shot generalization from a single training scene.}
	\label{fig:2d_paths}
\end{figure*}

In the 2D and 6D obstacle avoidance settings, $\mathcal{L}_{\text{pos}}$ is the mean cost of collision-free paths and $\mathcal{L}_{\text{neg}}$ is the mean cost of obstacle-penetrating paths. The ratio loss drives the field to make obstacle penetration expensive and free-space traversal cheap. In the 3D/4D black hole setting, $\mathcal{L}_{\text{pos}}$ is the mean cost of ingoing (infalling) paths and $\mathcal{L}_{\text{neg}}$ is the mean cost of outgoing (escaping) paths. A path is ``ingoing'' if it starts at a radius $r > r_{\text{max}}/2$ and ends at $r < r_{\text{max}}/4$, where $r_{\text{max}}$ is the radius of the spatial grid; ``outgoing'' if the reverse. The constraint is purely causal: it is easier to fall in than to climb out. No mass, no curvature, no Einstein equations are provided; only the relative cost of ingoing versus outgoing geodesics.

This ratio loss, however, admits a degenerate solution: the optimizer can shrink all metric eigenvalues toward zero, making $\mathcal{L}_{\text{pos}}$ and $\mathcal{L}_{\text{neg}}$ both arbitrarily small while maintaining a small ratio, without learning any geometric structure. We prevent this by enforcing $t_0 > 0$ on the coefficient of the identity basis matrix $X_0 = I$:
\begin{equation}
    t_0 = \mathrm{softplus}(t_0^{\mathrm{raw}}) + 10^{-6},
    \label{eq:softplus}
\end{equation}
which guarantees $\mathrm{tr}(H) = n \cdot t_0 > 0$ pointwise (all other basis matrices $X_k$, $k \ge 1$, are traceless).

The mechanism follows from the identity $\det(\exp(M)) = \exp(\mathrm{tr}(M))$. Under the Cartan clamp, $\tilde{H} = c \cdot H$ with $c > 0$, so $\mathrm{tr}(\tilde{H}) = c \cdot \mathrm{tr}(H) > 0$. In both geometric settings, this yields the same key result:
\begin{equation}
    |\det(G)| = \exp(2\,\mathrm{tr}(\tilde{H})) > 1.
    \label{eq:det_G}
\end{equation}
For Riemannian geometry ($\eta = I$), $G = \exp(2\tilde{H})$, so $\det(G) = \exp(2\,\mathrm{tr}(\tilde{H}))$ directly. For Lorentzian geometry ($\eta = \mathrm{diag}(-1,+1,\ldots,+1)$),
\begin{equation}
    \det(G) = \det(E)^2 \det(\eta) = -\exp(2\,\mathrm{tr}(\tilde{H})),
\end{equation}
yielding the same absolute value. Since $|\det(G)| > 1$, the eigenvalues of $G$ cannot all shrink toward zero simultaneously, since their product (or absolute product) would otherwise approach zero. Thus the metric cannot be uniformly collapsed; the ratio loss is forced to work through genuine geometric structure (making $\mathcal{L}_{\text{neg}}$ large relative to $\mathcal{L}_{\text{pos}}$) rather than through metric shrinkage.

\subsection{The Cartan Spectral Clamp}
\label{sec:seesaw}

The Cartan clamp (Eq.~\ref{eq:cartan}) serves two purposes. Mathematically, it bounds the spectral radius of $H(x)$, preventing the matrix exponential from producing numerically singular metrics. Physically, it controls the maximum curvature the field can express: a larger $L$ permits stronger local variations; a smaller $L$ produces smoother fields. The clamp parameter functions as a principled geometric control, not a regularization coefficient that requires tuning against a loss, but a bound on the intrinsic curvature of the learned manifold.

In the 3D and 4D experiments, we introduce spatial variation in $L$. Let $r$ be the radial distance from the center. The clamp parameter becomes:
\begin{equation}
    L(r) = L_f + (L_c - L_f) \cdot \sigma\!\left(\frac{r_{\mathrm{tr}} - r}{w}\right),
    \label{eq:L_radial}
\end{equation}
where $L_c$ (center clamp) allows strong curvature near the origin, $L_f$ (far clamp) tightly constrains the far region, $\sigma$ is the sigmoid function, $r_{\mathrm{tr}}$ is the transition radius, and $w$ controls the sharpness of the decay. This radial profile enables the Cartan seesaw mechanism discussed in Section~\ref{sec:seesaw}. For 2D and 6D, uniform Cartan clamping ($L = \text{constant}$) is sufficient.

\section{2D: Zero-Shot Obstacle Avoidance}
\label{sec:2d}

\begin{figure*}[t]
	\centering
	\begin{subfigure}[b]{0.25\textwidth}
		\centering
		\includegraphics[width=\textwidth]{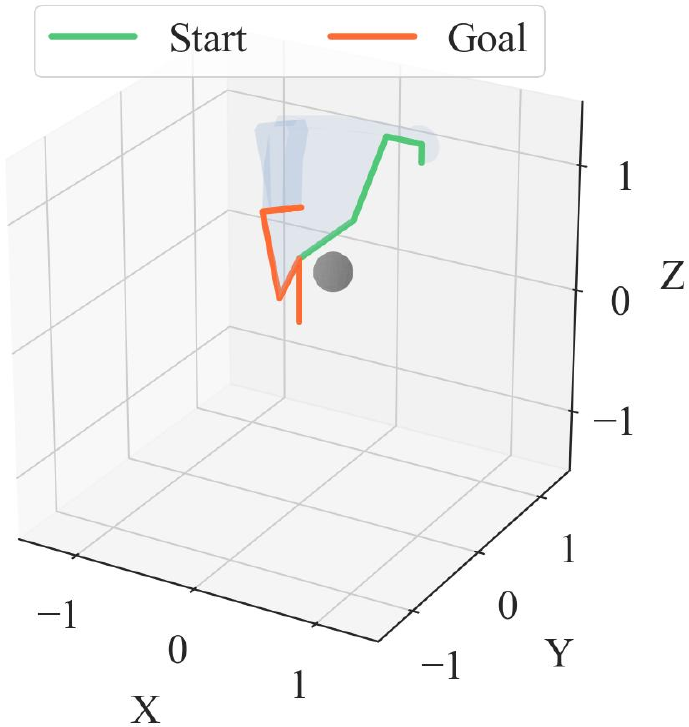}
		\caption{1 obstacle}
	\end{subfigure}\hfill
	\begin{subfigure}[b]{0.25\textwidth}
		\centering
		\includegraphics[width=\textwidth]{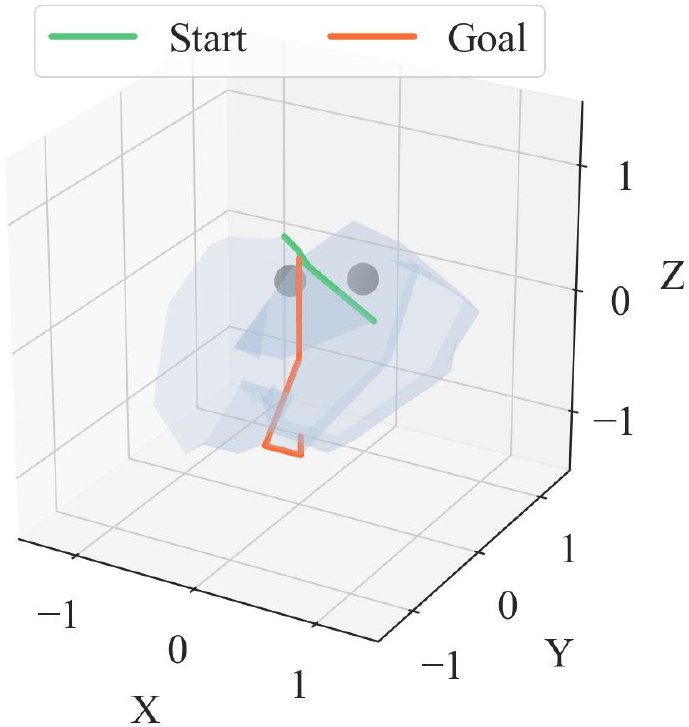}
		\caption{2 obstacles}
	\end{subfigure}\hfill
	\begin{subfigure}[b]{0.25\textwidth}
		\centering
		\includegraphics[width=\textwidth]{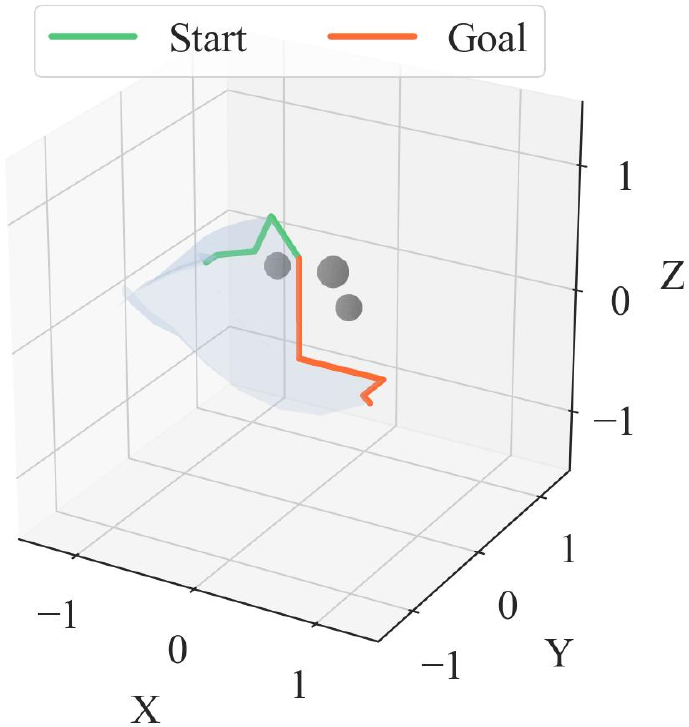}
		\caption{3 obstacles}
	\end{subfigure}\hfill
	\begin{subfigure}[b]{0.25\textwidth}
		\centering
		\includegraphics[width=\textwidth]{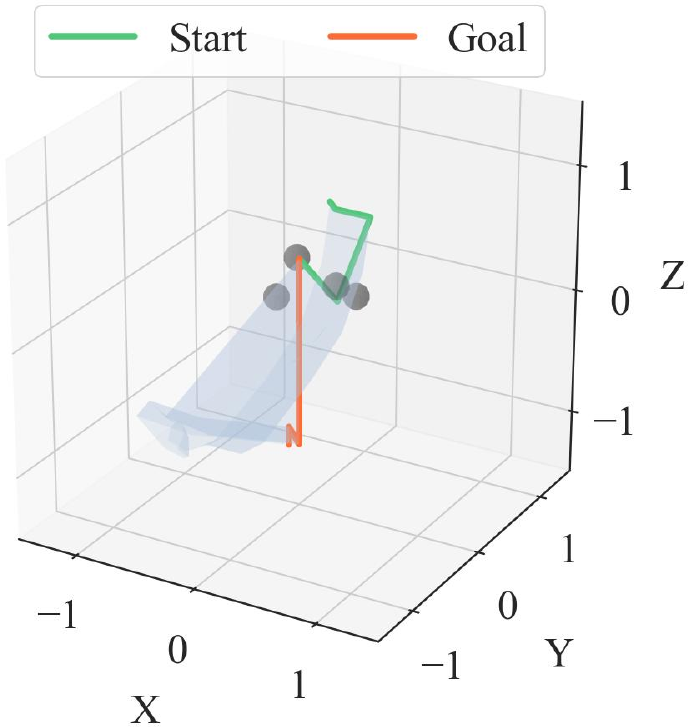}
		\caption{4 obstacles}
	\end{subfigure}
	\caption{6-DOF manipulator trajectories across four obstacle scenes.}
	\label{fig:6d_scene}
\end{figure*}

As a first validation, we apply the continuous metric field to 2D planar navigation. A point robot learns to avoid obstacles purely through metric geometry, with zero-shot generalization from a single training scene.

\subsection{Setup}

The 2D experiments use a planar point robot in a $3 \times 3$ world. The scene is rasterized into a $64 \times 64$ density grid. The metric is assembled via the unified form $G = E^2 = \exp(2\tilde{H})$ with $\eta = I$. Hyperparameters: Cartan clamp $L = 12$, learning rate $3\times10^{-3}$, 400 epochs, Adam optimizer, gradient clipping at 10.0. The $t_0 > 0$ constraint (Eq.~\ref{eq:softplus}) is applied to the coefficient of $X_0 = I$.

\subsection{Zero-Shot Generalization from a Single Training Scene}

We train on a single scene (T1): two circular obstacles of radius $0.3$ at $(0,0)$ and $(0.6,0.6)$. Start and goal positions are randomly sampled for each training path. The training pool consists of 600 collision-free (positive) and 1,800 obstacle-penetrating (negative) paths generated via A* planning and straight-line interpolation, respectively. A random subset of 80 positive and 160 negative paths is sampled per epoch. After training, we evaluate the learned metric field on a comprehensive zero-shot test suite spanning 5 difficulty levels:

\begin{itemize}
    \item \textbf{L1 (start/goal):} 10 scenes with the same obstacles but new start and goal positions.
    \item \textbf{L2 (position):} 10 scenes with obstacles shifted to new locations.
    \item \textbf{L3 (count):} 10 scenes with 1--7 obstacles.
    \item \textbf{L4 (shape):} 10 scenes with non-circular obstacles: rectangles, L-shapes, corridors.
    \item \textbf{L5 (random):} 50 fully randomized scenes.
\end{itemize}

All test scenes are filtered to guarantee that the straight-line path intersects at least one obstacle, ensuring non-trivial difficulty. Paths are extracted via A* search on the learned metric field, using only the metric cost $\sqrt{\delta^\top G \delta}$ with no hard collision checking or explicit obstacle representation.

The T1-trained model achieves 100\% pass rate on all 90 test scenes across all 5 levels. Figure~\ref{fig:2d_paths} shows representative paths on four challenging zero-shot test cases drawn from L4 (shape) and L5 (random). The learned metric field, visualized as local ellipses representing the inverse metric $G^{-1}(x)$, expands in obstacle regions and contracts in free space, guiding the A* geodesic to naturally avoid obstacles without any explicit collision constraint.

\section{6D: Manipulator Zero-Shot Generalization}
\label{sec:6d}

Having established that the metric field generalizes in 2D, we extend the architecture to a substantially harder domain: a 6-DOF serial manipulator operating in a 3D workspace with spherical obstacles. The 6-dimensional configuration space requires a $\mathrm{Sym}(6)$ metric, a 7$\times$ increase in parametric degrees of freedom over the 2D case.

\subsection{Setup}

The manipulator is defined by the standard Denavit-Hartenberg (DH) parameters in Table~\ref{tab:dh}. Forward kinematics are computed via the standard DH transformation, producing the 3D position of each joint from the 6D joint configuration $q \in [-\pi, \pi]^6$.

\begin{table}[t]
\centering
\caption{DH parameters of the 6-DOF manipulator.}
\label{tab:dh}
\begin{tabular}{@{}cccc@{}}
\toprule
Joint & $d$ (offset) & $a$ (link length) & $\alpha$ (twist) \\
\midrule
1 & 0.5 & 0.0 & $\pi/2$ \\
2 & 0.0 & 0.8 & 0 \\
3 & 0.0 & 0.7 & 0 \\
4 & 0.3 & 0.0 & $\pi/2$ \\
5 & 0.0 & 0.0 & $-\pi/2$ \\
6 & 0.15 & 0.0 & 0 \\
\bottomrule
\end{tabular}
\end{table}

The 3D workspace ($1.5 \times 1.5 \times 1.5$) is voxelized into a $32 \times 32 \times 32$ grid. Obstacles are spherical ($r \in [0.08, 0.18]$), placed randomly within the manipulator's reachable workspace. Collision checking is performed via forward kinematics at 20 interpolation points per path, with a link radius of 0.08, checking both joint positions and link segments against obstacle spheres. Paths are extracted via A* search on a discrete grid followed by geodesic optimization.

The metric field is parameterized by $\mathrm{Sym}(6)$ with $m=21$ basis matrices (6 diagonal, 15 off-diagonal). The 3D workspace voxel grid is encoded by a 3D CNN and mapped to the 21 coefficients $t_k(x)$. The Cartan clamp uses $L = 3.0$ uniformly. The $t_0 > 0$ constraint is applied to $X_0 = I$. Training uses 400 epochs, Adam with learning rate $3\times10^{-3}$, feature dimension $16$.

Paths are collision-free (pos) or collision-penetrating (neg) joint-space trajectories. The cost of a path is its Riemannian length in the learned 6D metric. The causal loss (Eq.~\ref{eq:loss}) drives the field to make collision-free paths cheap and colliding paths expensive.

\subsection{Zero-Shot Generalization from a Single Training Scene}

We train on a single scene (T1) containing two spherical obstacles in the workspace: one at $(0.07, 0.28, 0.29)$ with radius $0.16$, and the other at $(0.10, -0.09, -0.16)$ with radius $0.09$. The training pool consists of 19,200 paths (4,800 collision-free, 14,400 colliding), an 8$\times$ scale-up from the 2D T1 data volume. This larger pool is necessary to suppress overfitting in the 21-dimensional parameter space of $\mathrm{Sym}(6)$. Each training epoch randomly samples 80 positive and 160 negative paths from these pools. After training, we evaluate on a comprehensive zero-shot test suite spanning 6 difficulty levels across 91 total scenes, with 1,000 paths per scene for stable evaluation:

\begin{itemize}
    \item \textbf{L0 (baseline):} Identical to the training scene.
    \item \textbf{L1 (perturbation):} 10 scenes with obstacle positions perturbed by $\pm 0.15$.
    \item \textbf{L2 (position):} 10 scenes with obstacles regenerated at new random locations.
    \item \textbf{L3 (count):} 10 scenes with 1--6 randomly placed obstacles.
    \item \textbf{L4 (size):} 10 scenes with obstacle radii $r \in [0.05, 0.20]$.
    \item \textbf{L5 (random):} 50 fully randomized scenes (varying count, position, and size).
\end{itemize}

All test scenes are filtered to ensure the straight-line joint-space path intersects at least one obstacle. Paths are extracted via A* search on the learned 6D metric field, using only the metric cost $\sqrt{\delta^\top G \delta}$ with no hard collision checking.

The T1-trained model achieves 100\% pass rate on all 91 test scenes across all 6 levels, with a global median separation ratio of 3.07$\times$ and 89.0\% of scenes exceeding 2.0$\times$ (Table~\ref{tab:6d_t1}). The L0 baseline achieves 6.16$\times$, confirming that the learned metric generalizes well beyond the specific training paths. The separation ratio degrades gracefully under distribution shift, remaining above 2.7$\times$ median across all perturbation levels. This confirms that even in the 21-dimensional $\mathrm{Sym}(6)$ basis, the field captures transferable geometric structure from a single example. Figure~\ref{fig:6d_scene} visualizes the learned metric field in action: the A*-extracted geodesic guides the manipulator through 1--4 obstacles, with start and goal configurations in green and orange.

\begin{table}[t]
\centering
\caption{6D T1 zero-shot generalization results.}
\label{tab:6d_t1}
\begin{tabular}{@{}lccccccc@{}}
\toprule
& L0 & L1 & L2 & L3 & L4 & L5 & All \\
\midrule
Pass rate (\%)     & 100 & 100 & 100 & 100 & 100 & 100 & 100 \\
Median sep ($\times$) & 6.16 & 3.25 & 2.80 & 2.77 & 2.83 & 3.34 & 3.07 \\
Strong sep ($\ge 2\times$, \%) & 100 & 100 & 90 & 90 & 90 & 86 & 89 \\
\bottomrule
\end{tabular}
\end{table}

\section{3D: BTZ Black Hole Emergence}
\label{sec:3d}

Having established that the continuous metric field discovers obstacle geometry in 2D and generalizes zero-shot in 6D, we extend the architecture to 3D Lorentzian spacetime and ask: what does the causal contrast loss discover when the constraint is ``falling in is cheaper than climbing out''?

\subsection{Setup}

The 3D experiments use a $12 \times 12 \times 12$ spatial grid. The metric field is parameterized by $\mathrm{Sym}(3)$ with $m=6$ basis matrices. The Lorentzian basis is $\eta = \mathrm{diag}(-1, +1, +1)$. Ingoing paths start at $r > 0.5$ and end at $r < 0.25$ (pos); outgoing paths go the reverse direction (neg). Training uses 400 epochs, Adam with learning rate $3\times10^{-3}$, feature dimension $\text{fd}=20$, and 200 ingoing / 400 outgoing paths per batch.

The Cartan clamp uses the radial profile (Eq.~\ref{eq:L_radial}) with $L_c = 6.0$, $L_f = 0.5$, $r_{\mathrm{tr}} = 0.35$, and $w = 0.01$.

\subsection{Horizon Emergence}

In what follows, we denote components of the learned metric $G$ by their coordinate indices. $G_{tt}$ is the time-time component: negative in the exterior (time flows normally) and positive in the interior (time is trapped). $G_{xx}$, $G_{yy}$ are the spatial diagonal components, measuring proper distances along each spatial axis. The field discovers a black-hole-like metric in the optimal configuration. The defining signature of a black hole is the \emph{sign flip} of the time-time component $G_{tt}$: $G_{tt} < 0$ in the exterior (Lorentzian signature preserved) and $G_{tt} > 0$ in the interior. The radius where $G_{tt}$ crosses zero is the \emph{event horizon} $r_h$, a surface beyond which causal structure is trapped, where ingoing paths have significantly lower cost than outgoing paths. 

\begin{figure}[t]
\centering
\includegraphics[width=\columnwidth]{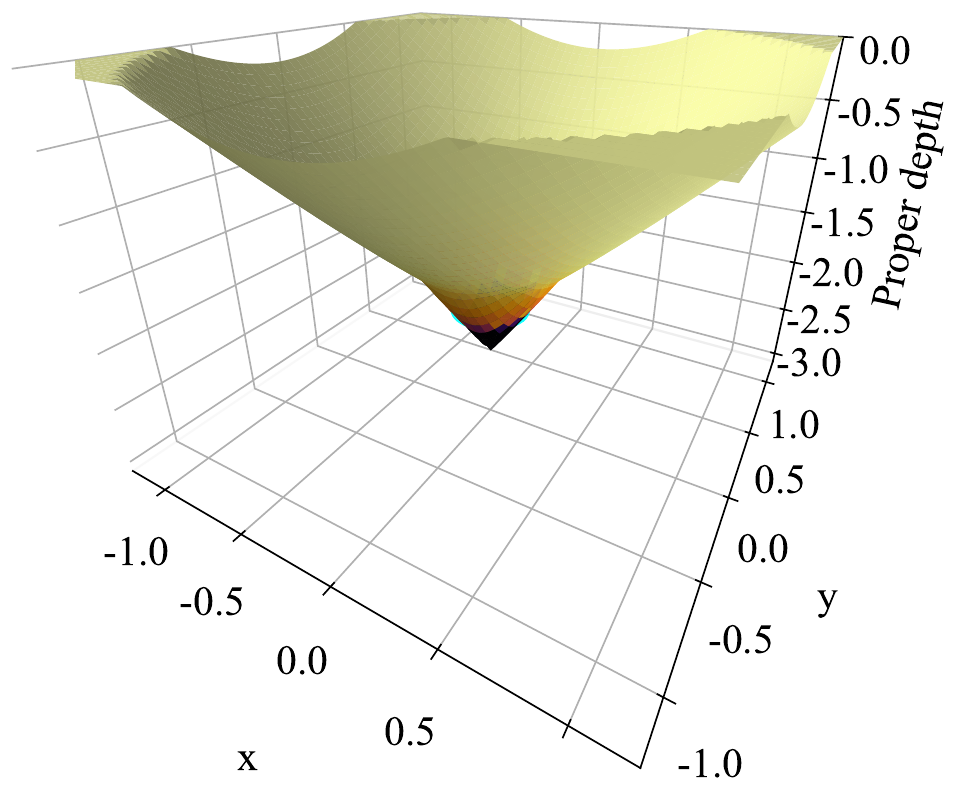}
\caption{Rubber-sheet visualization of the learned 3D black hole metric.}
\label{fig:3d_rubber}
\end{figure}

The learned field exhibits exactly this behavior: at the center $G_{tt}(0) = +0.189$ (positive, signature flipped), while in the far region $G_{tt} \approx -0.419$ (negative, Lorentzian preserved), with the sign flip occurring at $r_h \approx 0.184$ (Fig.~\ref{fig:3d_rubber}). The separation ratio between outgoing and ingoing path costs reaches $569\times$, demonstrating a strong causal asymmetry. The spatial components $G_{xx}$ and $G_{yy}$ remain positive definite. Crucially, this is a sign flip, a metric with one event horizon, analogous to the BTZ black hole.

\begin{figure}[t]
	\centering
	\includegraphics[width=0.85\columnwidth]{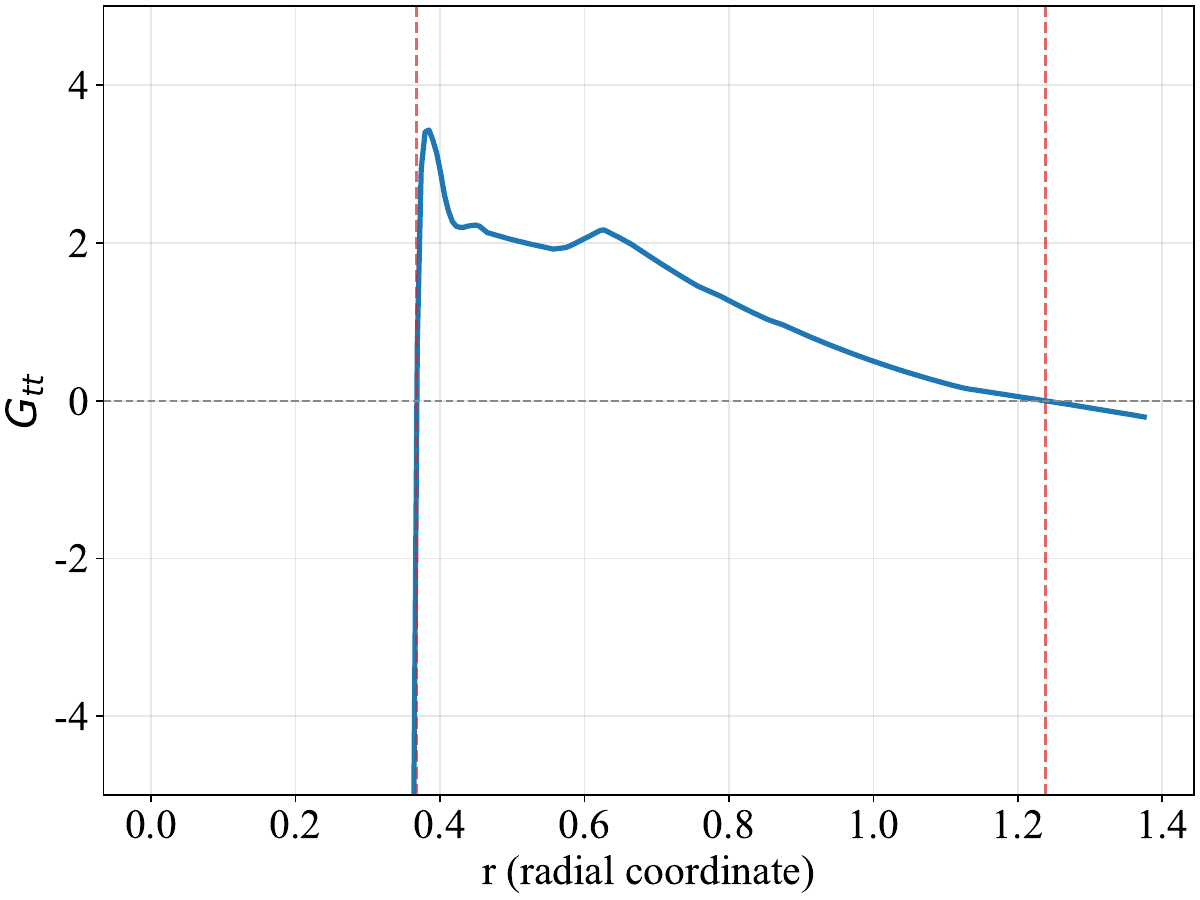}
	\caption{The shell flip: double sign change of $G_{tt}(r)$ (uniform $L=4.0$).}
	\label{fig:shell_flip}
\end{figure}

\subsection{Decay Parameter Scan}

The sharpness of the Cartan decay, controlled by the width $w$ in Eq.~\ref{eq:L_radial}, is the decisive parameter for horizon emergence. We scanned 12 configurations crossing four widths ($w \in \{0.08, 0.04, 0.02, 0.01\}$) and three transition radii ($r_{\mathrm{tr}} \in \{0.35, 0.20, 0.10\}$), with $L_c=6.0$, $L_f=0.5$ fixed.

\begin{table}[t]
\centering
\caption{3D decay parameter scan.}
\label{tab:decay_scan}
\begin{tabular}{@{}ccccccl@{}}
\toprule
$w$ & $r_{\mathrm{tr}}$ & $G_{tt}(0)$ & Sep & $r_h$ & BH? \\
\midrule
0.08 & 0.35 & $+0.291$ & 460$\times$ & 0.470 & \checkmark \\
0.08 & 0.20 & $-2.275$ & 159$\times$ & --- & \\
0.08 & 0.10 & $-14046$ & 49$\times$ & --- & \\
\midrule
0.04 & 0.35 & $-0.001$ & 573$\times$ & --- & \\
0.04 & 0.20 & $+0.013$ & 285$\times$ & 0.019 & \checkmark \\
0.04 & 0.10 & $-70621$ & 60$\times$ & --- & \\
\midrule
0.02 & 0.35 & $+0.243$ & 560$\times$ & 0.172 & \checkmark \\
0.02 & 0.20 & $+0.094$ & 352$\times$ & 0.207 & \checkmark \\
0.02 & 0.10 & $-151163$ & 89$\times$ & --- & \\
\midrule
0.01 & 0.35 & $+0.189$ & 569$\times$ & 0.184 & \checkmark \\
0.01 & 0.20 & $+1.402$ & 377$\times$ & 0.203 & \checkmark \\
0.01 & 0.10 & $-154778$ & 130$\times$ & --- & \\
\bottomrule
\end{tabular}
\end{table}

Three qualitative patterns emerge (Table~\ref{tab:decay_scan}). \textbf{First}, sharper decay promotes horizon formation: at $w=0.01$ and $w=0.02$, black-hole-like solutions (with a single sign flip) appear at both $r_{\mathrm{tr}}=0.35$ and $r_{\mathrm{tr}}=0.20$, while coarser decay ($w=0.04, 0.08$) yields at most one success. \textbf{Second}, $r_{\mathrm{tr}}=0.10$ universally fails regardless of $w$, producing deeply negative $G_{tt}$ at the center; the transition is too close to the origin, leaving insufficient room for the metric to develop curvature. \textbf{Third}, among the successful configurations, the two transition radii produce different regimes: $r_{\mathrm{tr}}=0.35$ consistently yields large separation (460--569$\times$) with a moderate $G_{tt}(0) \in [+0.19, +0.29]$, while $r_{\mathrm{tr}}=0.20$ yields more variable outcomes --- a weak horizon at coarse decay ($G_{tt}(0)=+0.013$) but an extreme one at sharp decay ($G_{tt}(0)=+1.402$), the latter approaching the shell-flip regime documented below. The sharp decay effectively decouples the center and far regions: the center enjoys full expressive freedom while the far region is tightly anchored, leaving the optimizer no choice but to produce a genuine sign flip.

\subsection{The Shell Flip}
\label{sec:shell_flip}

Under suboptimal Cartan configurations (insufficient $L_c$, or uniform clamping), a distinct failure mode appears: the \emph{shell flip}, where $G_{tt}$ changes sign \emph{twice} along the radial direction: negative in the far region, positive in an intermediate shell, and negative again at the center (Fig.~\ref{fig:shell_flip}). This corresponds to \emph{two} event horizons, with separation ratios of 9$\times$ (uniform $L=5.0$) and 152$\times$ (uniform $L=4.0$). The shell flip can appear under uniform clamping and is suppressed by the Cartan seesaw (Section~\ref{sec:seesaw}).

\begin{figure*}[t]
	\centering
	\begin{subfigure}[b]{0.32\textwidth}
		\centering
		\includegraphics[width=\textwidth]{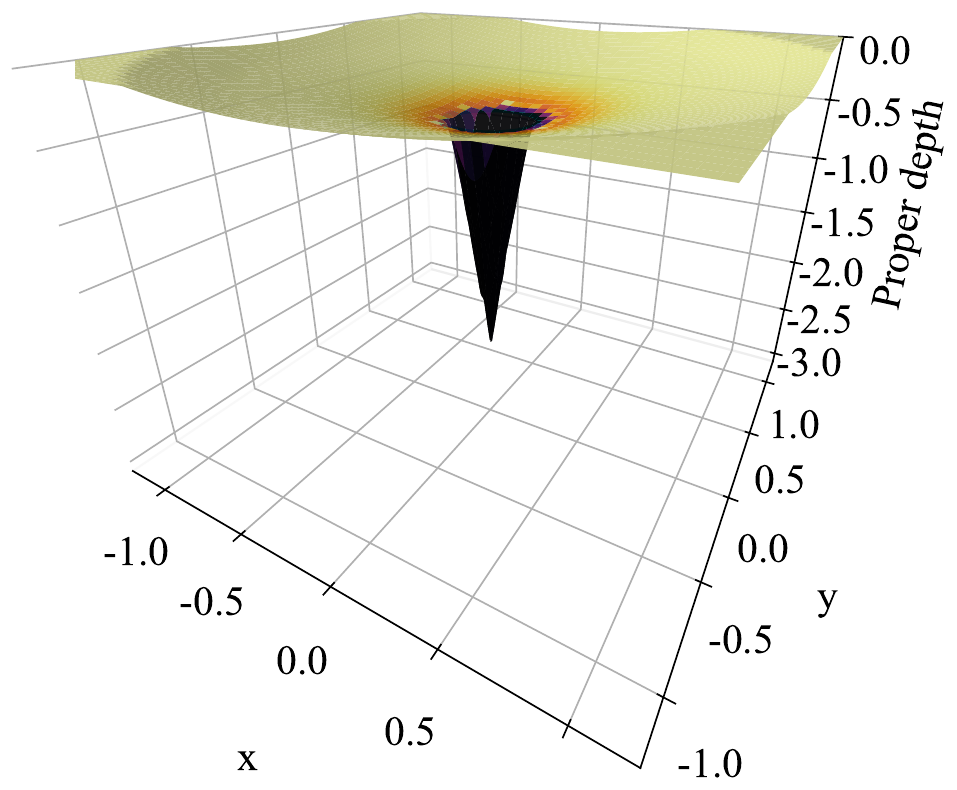}
		\caption{xy slice}
	\end{subfigure}\hfill
	\begin{subfigure}[b]{0.32\textwidth}
		\centering
		\includegraphics[width=\textwidth]{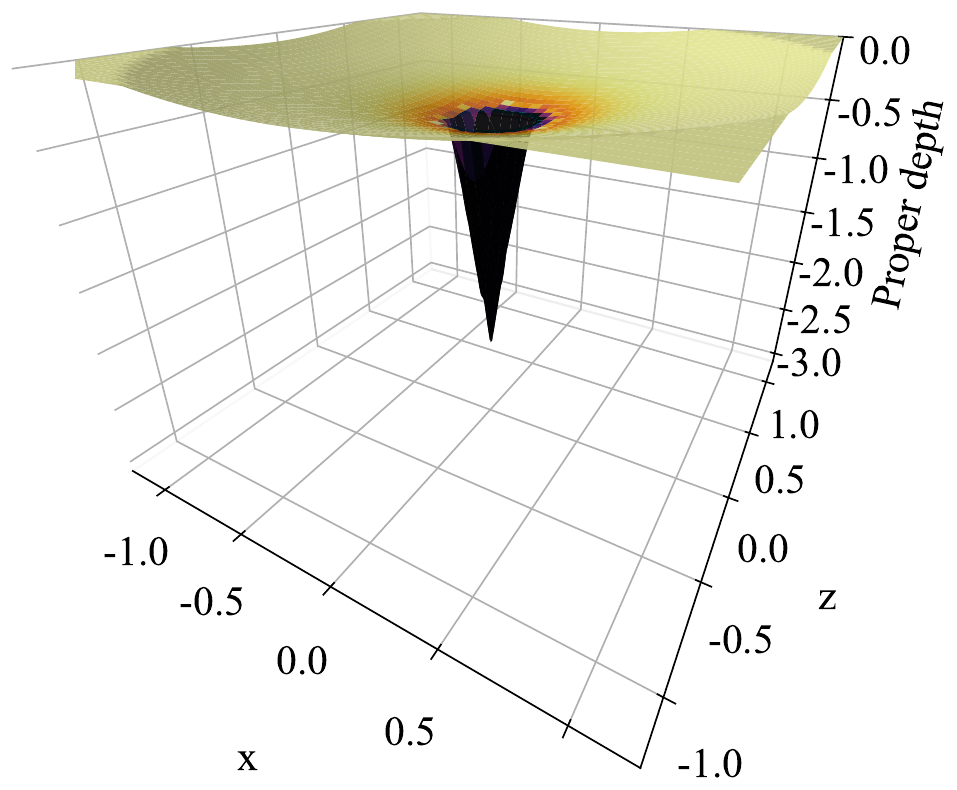}
		\caption{xz slice}
	\end{subfigure}\hfill
	\begin{subfigure}[b]{0.32\textwidth}
		\centering
		\includegraphics[width=\textwidth]{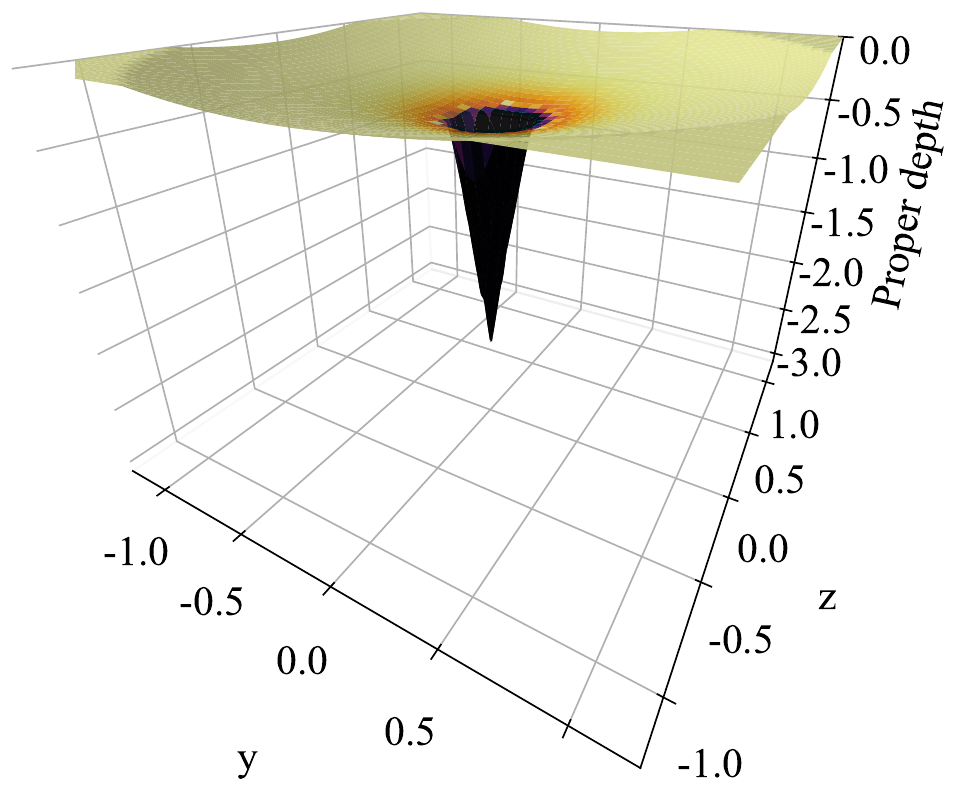}
		\caption{yz slice}
	\end{subfigure}
	\caption{Rubber-sheet visualization of the learned 4D black hole metric.}
	\label{fig:4d_rubber}
\end{figure*}
\section{4D: Schwarzschild Black Hole}
\label{sec:4d}
The 3D experiments demonstrate that a single sign flip, i.e., a black hole horizon, emerges from causal contrast alone. We now ask whether this phenomenon persists in higher dimensions. Extending to 4D Lorentzian spacetime tests whether the seesaw mechanism generalizes and whether the learned metric exhibits the symmetries expected of a physical black hole.

\subsection{Setup}

The 4D experiments extend the architecture to $\mathrm{Sym}(4)$ with $m=10$ basis matrices. The grid is $12 \times 12 \times 12 \times 12$. The Lorentzian basis is $\eta = \mathrm{diag}(-1, +1, +1, +1)$. Training uses feature dimension $\text{fd} = 20$, 400 epochs, Adam with learning rate $3\times10^{-3}$, and 200 ingoing / 400 outgoing paths per batch. The optimal configuration for results reported below uses the radial profile (Eq.~\ref{eq:L_radial}) with $L_c = 8.0$, $L_f = 0.5$, $r_{\mathrm{tr}} = 0.20$, and $w = 0.01$.

\subsection{Schwarzschild Emergence}

For physical interpretation, we report metric components on the underlying Cartesian grid. $G_{tt}$ retains the same meaning as in 3D---its sign flip marks the horizon. The off-diagonal components $G_{tx}, G_{ty}, G_{tz}$ are time-space cross terms; their smallness indicates that the field is approximately static and spherically symmetric.

The 4D field discovers a Schwarzschild-like black hole with separation ratio 628$\times$, characterized by two signatures. First, an event horizon: $G_{tt}$ is positive at the center ($+96.5$) and negative in the far region ($-0.536$), crossing zero at $r_h = 0.211$; inside $r_h$ the time direction is trapped, outside it is free, consistent with the sign pattern of the Schwarzschild metric. Second, emergent spherical symmetry: the three off-diagonal time-space components ($G_{tx}, G_{ty}, G_{tz}$) are all suppressed below $4\times10^{-2}$ without any symmetry constraint, indicating that the $m=10$ basis provides sufficient independent control to suppress cross-terms while maintaining the $G_{tt}$ sign flip. The spatial diagonal components remain close to unity ($\approx 1.04$) at the far region, close to the Minkowski value.

\subsection{$L_c$ Scan}

\begin{table}[t]
\centering
\caption{4D $L_c$ scan.}
\label{tab:4d_lscan}
\begin{tabular}{@{}lccc@{}}
\toprule
$L_c$ & $G_{tt}(0)$ & $r_h$ & Sep \\
\midrule
3.0 & $-282$ & --- & 21$\times$ \\
4.0 & $-124$ & --- & 50$\times$ \\
5.0 & $-2337$ & --- & 96$\times$ \\
6.0 & $-59685$ & --- & 219$\times$ \\
7.0 & $+0.017$ & 0.204 & 281$\times$ \\
8.0 & $+72.3$ & 0.211 & 623$\times$ \\
\bottomrule
\end{tabular}
\end{table}

Table~\ref{tab:4d_lscan} shows the effect of increasing $L_c$ while keeping $L_f=0.5$. For $L_c \leq 6.0$, $G_{tt}(0)$ is deeply negative, and the center remains causally ``free'', with no horizon forming. At $L_c = 7.0$, $G_{tt}(0)$ barely crosses zero ($+0.017$), producing a weak horizon. At $L_c = 8.0$, the center becomes strongly positive ($+72.3$), producing a robust horizon with 623$\times$ separation. The seesaw requires a sufficiently large $L_c$ to overcome the anchoring effect of $L_f=0.5$.

\section{Conclusion}
\label{sec:conclusion}

We have demonstrated that a continuous metric field, constructed via a fixed basis of symmetric matrices and trained by a single causal contrastive loss, discovers the full cross-dimensional spectrum of geometric structures, from zero-shot obstacle avoidance in 2D and 6D, to BTZ-like horizons in 3D, to Schwarzschild-like black holes in 4D, using only causal signals. The same loss function and assembly principle produce this range of phenomena across dimensions. The broader implication is that causal principles alone suffice to drive the emergence of diverse geometric structures: a geometry satisfying ``good paths are cheaper than bad paths'' will, under sufficient parametric freedom, develop the geometric structures appropriate to its dimension.

\bibliographystyle{IEEEtran}
\bibliography{paper2}

\end{document}